\documentclass[conference]{IEEEtran}
\usepackage{amssymb}
\usepackage{bm}
\usepackage{textcomp}
\usepackage{cite}
\usepackage{CJK}
\usepackage{booktabs}
\usepackage{verbatim}
\usepackage{amsfonts}
\usepackage[top=2cm, bottom=2cm, left=2cm, right=2cm]{geometry}
\usepackage{algorithm}
\usepackage{algorithmicx}
\usepackage{algpseudocode}
\usepackage{amsmath}
\usepackage{graphicx}
\usepackage{subfigure}
\usepackage{multirow}
\usepackage{verbatim}
\newcommand{\bz}{\boldsymbol{z}}
\newcommand{\bx}{\boldsymbol{x}}

\makeatletter
\newcommand*\bigcdot{\mathpalette\bigcdot@{.5}}
\newcommand*\bigcdot@[2]{\mathbin{\vcenter{\hbox{\scalebox{#2}{$\m@th#1\bullet$}}}}}
\makeatother
\begin{document}
\title{Federated Traffic Synthesizing and Classification Using Generative Adversarial Networks}
\author{\IEEEauthorblockA{Chenxin~Xu\IEEEauthorrefmark{1}, Rong~Xia\IEEEauthorrefmark{1}, Yong~Xiao\IEEEauthorrefmark{1}\IEEEauthorrefmark{3},Yingyu~Li\IEEEauthorrefmark{1}, Guangming~Shi\IEEEauthorrefmark{2}\IEEEauthorrefmark{3}, and Kwang-cheng Chen\IEEEauthorrefmark{4} \\
\IEEEauthorblockA{\IEEEauthorrefmark{1}School of Electronic Inform. \& Commun., Huazhong Univ. of Science \& Technology, China}
\IEEEauthorblockA{\IEEEauthorrefmark{2}School of Artificial Intelligence, Xidian University, Xi'an, China}
\IEEEauthorblockA{\IEEEauthorrefmark{3}Pazhou Lab, Guangzhou, China}
\IEEEauthorblockA{\IEEEauthorrefmark{4}Department of Electrical Engineering, University of South Florida, FL}
}
}
\maketitle

\begin{abstract}
With the fast growing demand on new services and applications as well as  the increasing awareness of data protection, traditional centralized traffic classification approaches are facing unprecedented challenges. This paper introduces a novel framework, Federated Generative Adversarial Networks and Automatic Classification (FGAN-AC),  which integrates decentralized data synthesizing with traffic classification. FGAN-AC is able to synthesize and classify multiple types of service data traffic from decentralized local datasets without requiring a large volume of manually labeled dataset or causing any data leakage. Two types of data synthesizing approaches have been proposed and compared: computation-efficient FGAN (FGAN-\uppercase\expandafter{\romannumeral1}) and communication-efficient FGAN (FGAN-\uppercase\expandafter{\romannumeral2}). The former only implements a single CNN model for processing each local dataset and the later only requires coordination of intermediate model training parameters. An automatic data classification and model updating framework has been proposed to automatically identify unknown traffic from the synthesized data samples and create new pseudo-labels for model training. Numerical results show that our proposed framework has the ability to synthesize highly mixed service data traffic and can significantly improve the traffic classification performance compared to existing solutions.
\end{abstract}
\vspace{-0.2cm}
\section{Introduction}
  Traffic classification and identification have been considered as fundamental for a wide range of applications such as network anomaly detection and identification, as well as resource slicing for different prioritized services and network access control. Traditional traffic classification \cite{Info12,Info13,Info19} approaches are mostly centralized data processing approaches based on either supervised learning which often relies on a large volume of high-quality labeled datasets to train a model or clustering which divides data samples based on some observable features.

With fast growing demands on innovative services and applications, the next generation telecommunication technology is expected to support a plethora of new services and applications\cite{XY2020SelfLearn,XY2018TactileInternet}. Traditional traffic classification solutions are facing unprecedented challenges. More specifically, with new services and applications being introduced more frequently than ever, it becomes more difficult to collect a sufficient number of high-quality labeled datasets for each emerging service in its early stage of development. Clustering-based approaches suffer from limited accuracy and often incapable of identifying new unknown services, especially when the total volume of traffic data is relatively low. Furthermore, with the increasing awareness of data protection and user privacy, centralized data processing solutions relying on datasets uploaded to a single computer server will soon be impossible to support future services with stringent privacy and data protection requirements.

Federated learning (FL) is an emerging machine learning framework that enables collaborative model training based on decentralized datasets \cite{fl1,fl2}. It has been considered as one of  key enabling technologies for data processing and model construction based on decentralized datasets distributed across a large networking system. In spite of merits, FL suffers from several known limitations. In particular, traditional FL (e.g. FedAvg \cite{fl2}) is still a supervised learning approach and requires a large number of labeled datasets for training the model. Moreover, FL is known to suffer from slow convergence rate especially when the datasets are mostly unlabeled and heterogeneous (e.g. non-IID).

In this paper, we focus on traffic synthesizing and classification across decentralized  datasets distributed throughout a large geographic area. A set of local fog servers, each can access a local dataset (e.g. dataset collected in a local sub-region), have been deployed for training a global traffic classification model without requiring any transferring or data leakage of local datasets. We propose Federated Generative Adversarial Networks and Automatic Classification (FGAN-AC), a novel decentralized traffic synthesizing and classification framework. FGAN-AC seamlessly integrate a federated generative learning model called FGAN into an automatic classification framework. In FGAN, two types of generative learning algorithms have been introduced: computation-efficient FGAN (FGAN-\uppercase\expandafter{\romannumeral1}) and communication-efficient FGAN (FGAN-\uppercase\expandafter{\romannumeral2}). The former algorithm FGAN-\uppercase\expandafter{\romannumeral1} deploys a single model (discriminator) at each fog server to compare the synthetic data generated by the global generator with its local dataset. The later FGAN-\uppercase\expandafter{\romannumeral2} algorithm deploys a complete Generative Adversarial Networks (GANs) \cite{GAN} consisting of two models, a generator and a discriminator, at each fog server. The fog servers will then coordinate their model training via intermediate  parameters exchanging which usually require much less communication overhead compared to exchanging real or fake (synthetic) data. We show that both generative learning algorithms create synthetic samples that capture the mixture of distribution of all the decentralized datasets. Based on these synthetic data samples, an automatic classification and model updating  framework has been proposed to automatically identify unknown traffic and assign pseudo-labels for model training. Extensive simulations have been conducted to evaluate the performance of the proposed framework. Our results show that FGAN-AC is able to synthesize highly mixed service data traffic and can significantly improve the traffic classification performance.

\vspace{-0.25cm}
\section{Related Work}
\vspace{-0.15cm}
\noindent{\bf Federated Learning:}
  Federated learning is an emerging distributed machine learning solution enabling collaborative model training based on decentralized datasets \cite{MD27,MD41,Fs28,Fs29}. It has been proved to be an very effective solution for reducing communication overhead \cite{MD27} and increasing data privacy \cite{MD41}. Recent results suggest that combining FL and generative learning is possible to create synthetic data samples that capture the mixture of distributions associated with decentralized datasets \cite{MD,Fed,MD42}. For example, in \cite{Fed}, the authors proposed a framework that could apply multiple  discriminators and generators to synthesize dataset with complex features.\\
\noindent{\bf Deep Learning Based  Traffic Classification:}
  Deep neural network is considered as a promising solution for traffic classification \cite{Info12,Info13,Info19,Info20,Info21}. In particular, the  authors in \cite{Info21} introduced a novel solution with attention mechanism that extracted features of the payload segments and output the classification results through Softmax classification layer. 
   The authors in \cite{Info} proposed an classification model that can identify unknown classes of data and then automatically updating its model. Different from the above works, we integrate federated generative learning into a automatic classification framework to create synthetic samples to assist the automatic traffic classification. To the best of our knowledge, this is the first work that exploits data synthesizing for improving automatic traffic  classification performance.
\vspace{-0.3cm}
\section{System Model And Architecture Overview}
\vspace{-0.2cm}
\subsection{System Model}
\vspace{-0.2cm}
  We consider a fog computing networking system that can access and process a set of decentralized datasets. In particular, the system consists of the following elements:

\textbf{Local datasets} consist of  service data generated by users in different service coverage areas. We assume there are no overlapping among different datasets and each dataset consists of a limited  number of types of service traffic.

 \textbf{Local fog servers} are deployed at different service areas to process its local datasets. We assume each fog server is associated with a local dataset. Due to the privacy consideration, each dataset can only be accessed by the associated fog server.

 \textbf{Global coordinator} is the computing server connected to local fog servers across all service areas. The coordinator cannot access any local dataset, but can communicate with all the fog servers and process their uploaded intermediate results.
\vspace{-0.2cm}
\subsection{Architecture Overview}
\vspace{-0.2cm}
 The main objective is to construct a joint training model based on the decentralized datasets to synthesize and classify different service traffic data. The joint training process will be coordinated by the global coordinator. We propose a novel framework called FGAN-AC that  can capture the distributions of multiple decentralized datasets and perform global service synthesizing and classification without resulting any data leakage. In the FGAN-AC approach, we propose a three-step framework for automatic data synthesizing and classification described as follows:
\begin{figure}[htb]
\centering
\includegraphics[width=8.0cm,scale=1.5]{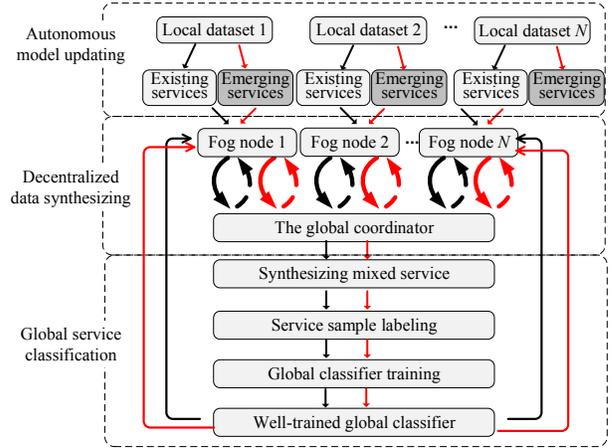}
\vspace{-0.15in}
\caption{Architecture overview}
\vspace{-0.20in}
\end{figure}

\noindent {\bf Decentralized Service Synthesizing:} In this step, a large volume of synthetic service traffic will be created by the global coordinator to capture the mixed distributions of traffic data from all local datasets. We borrow the idea from federated learning to efficiently aggregate the models deployed and trained over different fog servers. Two types of FGAN approaches are proposed to coordinate among fog servers as shown in Fig. 2: computation-efficient FGAN, labeled as FGAN-\uppercase\expandafter{\romannumeral1}, and communication-efficient FGAN, labeled as FGAN-\uppercase\expandafter{\romannumeral2}.
\begin{figure}[htbp]
\centering
\subfigure[FGAN-\uppercase\expandafter{\romannumeral1}]{
\begin{minipage}[t]{0.5\linewidth}
\centering
\includegraphics[width=3.8cm]{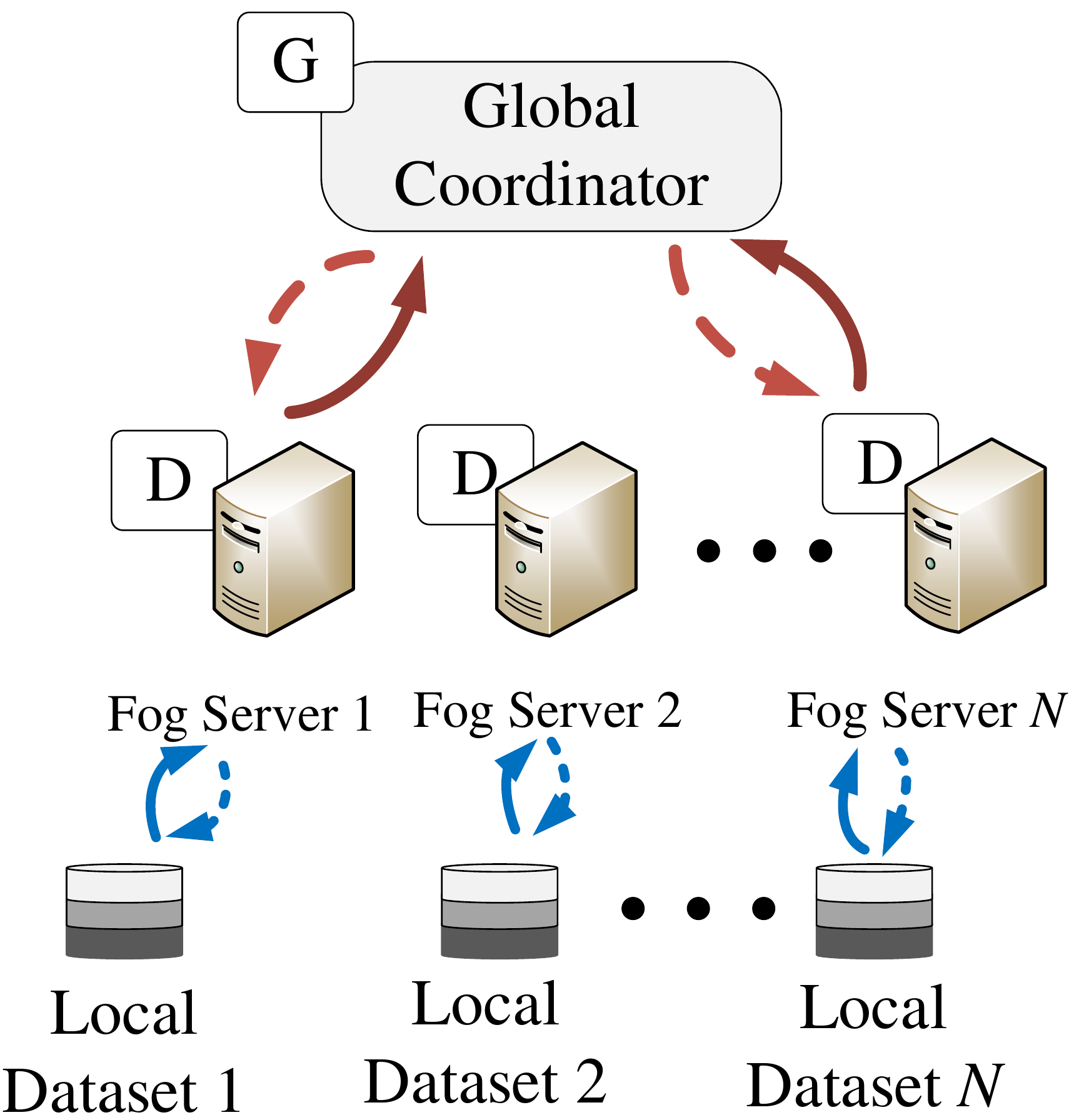}
\vspace{-0.6in}
\label{2(a)}
\end{minipage}%
}%
\subfigure[FGAN-\uppercase\expandafter{\romannumeral2}]{
\begin{minipage}[t]{0.5\linewidth}
\centering
\includegraphics[width=3.8cm]{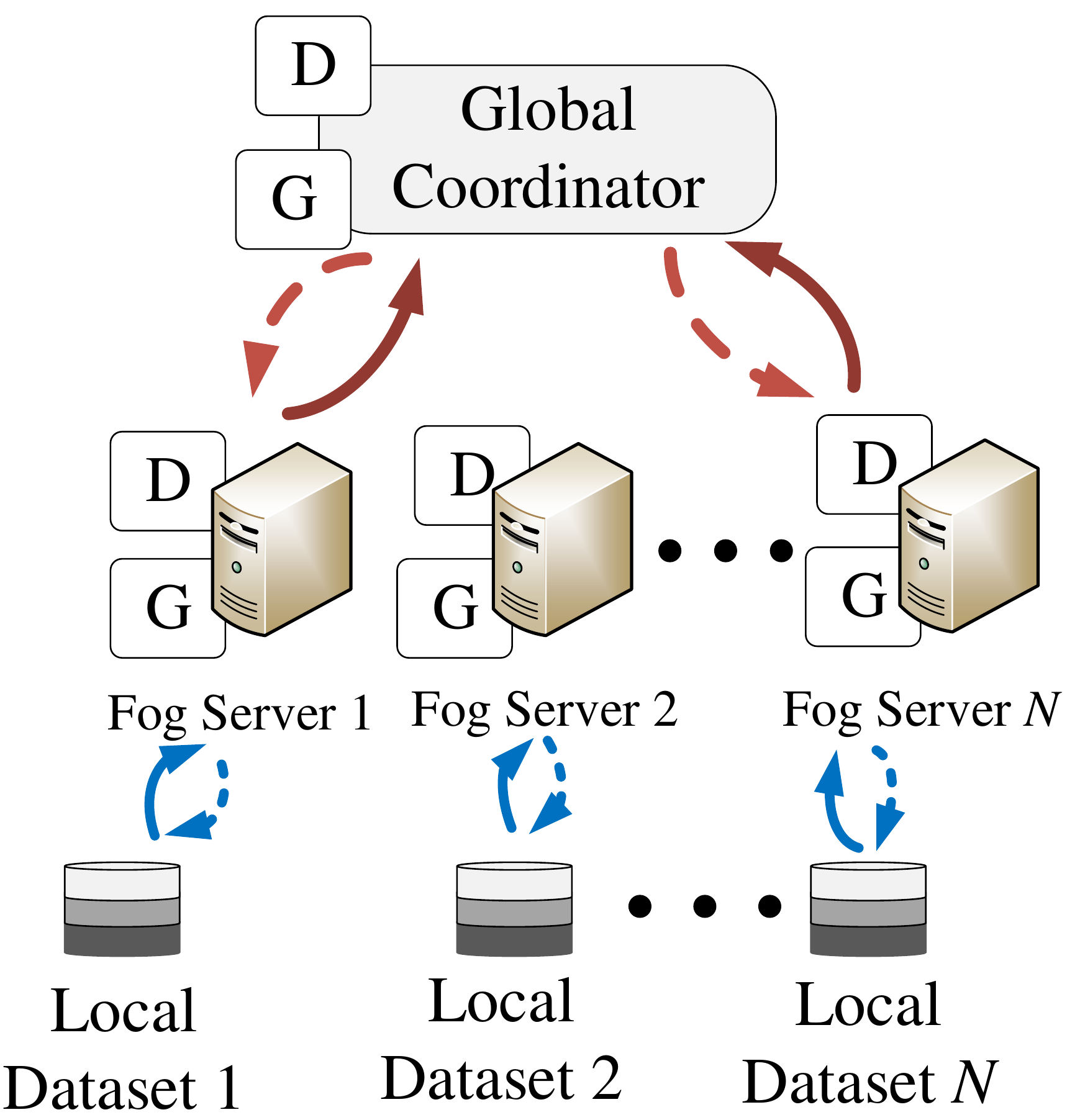}\vspace{-0.1in}
\label{2(b)}
\end{minipage}%
}%
\centering
\vspace{-0.15in}
\caption{FGAN Architecture.}
\end{figure}

{\it FGAN-I:} In this approach, only a single model  is deployed at each fog server and the coordinator. More specifically, the coordinator will be implemented with a generator to create synthetic data to be transmitted to each fog server where a local discriminator is deployed to distinguish between the real traffic data and the synthetic data received from the generator. The generator in the coordinator will be compete with all the discriminators in fog servers to generate a mixture of data samples related to the services in local datasets.

{\it FGAN-II:} In this approach, a GANs model consisting of both a generator and a discriminator will be deployed at each fog server as well as the coordinator. In this case, all fog servers will coordinate their intermediate model training parameters with each other via the coordinator. The coordinator will first aggregate all the uploaded parameters and then broadcast the updated ones to each fog server.

\noindent{\bf Global Service Classification:} In this step, a global service classification model will be trained based on  the synthesized dataset. In particular, we first adopt a deep clustering approach, named as Deep Embedded Clustering (DEC) \cite{DEC}, to create pseudo-labels for high-quality synthesized service data samples. Different from traditional clustering algorithms which heavily rely on the original dataset, DEC further exploits the latent features of the  input data to help separate different data traffic. Secondly, we feed the above pseudo-labeled dataset to a classifier located at the coordinator for supervised model training. Finally,  the global coordinator will broadcast the well-trained classification model to all the connected fog servers for local data classification and unknown traffic filtering. Trained with global mixed service samples, the broadcasted classification model can not only distinguish local dataset for each individual fog server, but also identify the potential service data that only available in other datasets.

\noindent{\bf Autonomous Model Updating:} In this step, a self-updating mechanism is adopted to autonomously identify newly arrived  unknown service traffics and assign new pseudo-labels for them. And then those newly synthesized service samples will be merged with the original dataset to re-train the classifier. After the training process is completed, the global coordinator will broadcast the updated classifier again for future traffic identification.

\section{FGAN-AC Architecture}
  In this section, we present more detailed descriptions of the three main steps of FGAN-AC.
\subsection{Decentralized Service Synthesizing}
  As mentioned earlier, one main objective of this paper is to synthesize sufficient service traffic samples  that can capture the distribution of the real service datasets located in multiple fog servers. In particular, we utilize a popular deep generative neural network, GANs, to produce high-quality samples. GANs consist of two neural networks referred to as generator $G$ and discriminator $D$ that competitively play a two player min-max game. The generator tries to fool the discriminator with synthetic samples mapped from Gaussian noises, and the discriminator aims to differentiate the real data samples from the fake ones. The payoff value of GANs can  be formulated as:
\begin{equation} \small \label{1}
\begin{aligned}
\min_G\max_D V(G,D)=\mathbb{E}_{\bx\sim P_{data}}[\log D(\bx)] \\+\mathbb{E}_{\bz\sim P_{z}}[\log\left(1-D(G(\bz))\right)],
\end{aligned}
\end{equation}
where $\bx$ denotes data samples  drawn from real dataset, $P_{data}$ is the data distribution of real dataset, and $\bz$ is drawn from Gaussian space $P_z$. In this step, we train a decentralized GANs model over multiple datasets to efficiently capture the distribution of mixed service traffic collected by different fog servers. In particular, we  introduce two types of FGAN approaches: computation-efficient FGAN-\uppercase\expandafter{\romannumeral1} and communication-efficient FGAN-\uppercase\expandafter{\romannumeral2}.
\subsubsection{FGAN-\uppercase\expandafter{\romannumeral1}}
As shown in Fig. \ref{2(a)}, in FGAN-\uppercase\expandafter{\romannumeral1},  a global generator $G$ located at the coordinator is competing with $N$ decentralized discriminator $\{D_1, D_2, \dots, D_N\}$, each associated with an exclusive dataset $d_n$. We can write the pay-off value of FGAN-\uppercase\expandafter{\romannumeral1} as follows:
\begin{subequations}\small \label{2}
\begin{align}
 \min_G V(G)&=\frac{1}{N}\sum_{n=1}^{N}\mathbb{E}_{\bz\sim P_{z}}[\log\left(1-D_n(G(\bz))\right)], \\
\max_{D_n} V(D_n)&=\mathbb{E}_{\bx\sim P_{d_n}}[\log D_n(\bx)]\nonumber \\
                  &+\mathbb{E}_{\bz\sim P_{z}}[\log\left(1-D_n(G(\bz))\right)],
\end{align}
\end{subequations}
 where $P_{d_n}$ is the data distribution of real dataset $d_n$. In each joint training round, $G$ first generates two batches of fake samples with batch size $b$, denoted as $\bx^{(d)}$ and $\bx^{(g)}$, and sends them to all the connected fog servers. $\bx^{(d)}$ is used for training the discriminator, and $\bx^{(g)}$ is used for the discriminator to calculate loss value of the generator. Secondly, each discriminator $D_n$ deployed on a fog server draws a batch of samples  $\bx^{(r)}$  from its local dataset $d_n$ and performs a $E$-step updating by ascending the gradient of loss value $L_{disc}^{n}$. Thirdly, each updated discriminator $D_n$ will utilize $\bx^{(g)}$ to calculate a loss value $L_{gen}^n$ for the generator. After receiving all the feedbacks from $N$ fog servers, the global generator $G$ will also update by descending the gradient of an aggregated loss value $L_{gen}$. We summarize the training process of FGAN-\uppercase\expandafter{\romannumeral1} in Algorithm 1. The above mentioned loss functions for model updating are formulated as follows:
\begin{subequations}\small \label{3}
\begin{align}
\mathop L\nolimits_{disc}^{n}&=\frac{1}{b}\sum\limits_{x \in \mathop {\bx} \nolimits^{(r)} } {\log (\mathop D\nolimits_n (x)) + \frac{1}{b}} \sum\limits_{x \in \mathop {\bx}\nolimits^{(d)} } {\log (1 - \mathop D\nolimits_n(x))}, \\
L_{gen}^n &= \frac{1}{b}\sum\limits_{x \in \mathop {\bx}\nolimits^{(g)}} {\log (1 -  D_n (x))}, \\
 L_{gen}  &= \frac{1}{N}\sum\limits_{n = 1}^N { L_{gen}^n}.
\end{align}
\end{subequations}
\begin{CJK}{UTF8}{gkai}
    \begin{algorithm}
\scriptsize
        \caption{FGAN-\uppercase\expandafter{\romannumeral1}}
            {\bf Input:} maximum global training round $I$, local training epoch $E$;\\
            {\bf Output:} global generator $G$ ;\\
				{\bf  for} global training round $i\leq I$  {\bf  do}
             \begin{itemize}
              \item[] Global generator broadcasts two batches of synthesized samples ${\bx}^{(g)}$ and ${\bx}^{(d)}$;
              \item[] {\bf for} each local discriminator {\bf parallel do}
              \item[] \quad $L_{gen}^n \gets {LocalUpdate}{({\bx}^{(g)},{\bx}^{(d)})}$
              \item[] {\bf end for}
					\item[] Aggregate local loss values by  Equation (3c)
					  \item[] Update generator by descending the gradient of the aggregated loss;
          	\end{itemize}
           {\bf end for}

               {\bf Function}\ {$LocalUpdate$} ${({\bx}^{(g)},{\bx}^{(d)})}$\\
              \quad {\bf for} local training round $e\leq E$  {\bf do}
\begin{itemize}
                 \item[]  1)~Sample a batch of real data  ${\bx}^{(r)}$;
                 \item[]  2)~Update discriminator by ascending the gradient of Equation (3a);
							\item[]  3)~Calculate the loss value of global generator by Equation (3b);
							\item[]  4)~ Upload the loss value to global generator;
\end{itemize}
 {\bf end for} \\
                \Return {loss value of global generator}

    \end{algorithm}
\end{CJK}

In FGAN-\uppercase\expandafter{\romannumeral1}, only a single component of GANs is  deployed at each fog server and coordinator, which significantly reduces the computation complexity. However, a large amount of synthesized data samples need to be transmitted between the global coordinator and each fog server in the training process. When the batch size is large or the data sample is complex which is common in practice, the communication overhead of FGAN-\uppercase\expandafter{\romannumeral1} will be extremely heavy.

\subsubsection{FGAN-\uppercase\expandafter{\romannumeral2}}
  In order to avoid redundant data sample transmission, in this approach, both the  generator and the discriminator are deployed at each fog server as well as the global coordinator as shown in Fig.~\ref{2(b)}. Different from FGAN-\uppercase\expandafter{\romannumeral1}, we borrow the idea from federated learning \cite{fl2} to efficiently aggregate information obtained from multiple decentralized datasets.
\tiny
\begin{CJK}{UTF8}{gkai}
    \begin{algorithm}
\scriptsize
        \caption{FGAN-\uppercase\expandafter{\romannumeral2}}
            {\bf Input:}  local GANs with parameters $w_n$ and $\theta_n$,   global GANs with parameters $w$ and $\theta$, global training round $I$, number of local training epoch $E$; \\
             {\bf Output:} $w$ and $\theta$; \\
       \quad {\bf  for} global training round $i\leq I$  {\bf  do}
       \begin{itemize}
       \item[] {\bf for} each local GANs {\bf parallel do}
         \item[] \quad $(w_{n},\theta_{n}) \gets {LocalUpdate}{(w,\theta)}$;
       \item[] {\bf end for}
       \item[] Aggregate received parameters by Equation  \eqref{9};
		\end{itemize}
    {\bf end for}\\
    \Return$w$,$\theta$

    {\bf  Function}\ {LocalUpdate} {($w,\theta$)}\\
     $w_n \gets w$, $\theta_n \gets \theta$;\\
     {\bf  for} i=1 $\to$ E {\bf  do}
\begin{itemize}
     \item[] Update $\theta_n$ by ascending the gradient of Equation (5a);
     \item[] Update $w_n$  by descending the gradient of Equation (5b);
    \end{itemize}
{\bf  end for}\\
    \Return{$w_{n},\theta_{n}$}

    \end{algorithm}
\end{CJK}
 \normalsize
 Firstly, in each global training round, the global GANs model located at the coordinator broadcasts its latest model parameters, denoted as $w$ for generator and $\theta$ for discriminator, to each fog server. Secondly, local GANs models located at different fog servers update their parameters  according to the received $w$ and $\theta$, and then performs an $E$-step local model updating by ascending the gradient of loss function $L_{D}^n$ and descending the gradient of loss function $L_{G}^n$. After the local training process is completed, local parameters denoted as $w_n$ and $\theta_n$ will be report to the global coordinator. At last, the global coordinator aggregates all the received local parameters by:
 \vspace{-0.3cm}
\begin{equation}\small \label{9}
	\begin{aligned}
	w = \frac{1}{N}\sum_{n = 1}^N {\mathop w\nolimits_n } ,\quad
	\theta = \frac{1}{N}\sum_{n = 1}^N {\mathop \theta \nolimits_n }.
	\end{aligned}
	\vspace{-0.3cm}
\end{equation}
Following the notations in FGAN-\uppercase\expandafter{\romannumeral1}, $\bx^{(r)}$ is a batch of real service samples, $\bx^{(g)}$ and $\bx^{(d)}$ denotes a batch of synthesized samples used for training generator and discriminator, respectively. The loss functions of the discriminator and the generator at each fog server $n$ can be formulated as:
\vspace{-0.1cm}
\begin{subequations} \small \label{7}
\begin{align}
&\mathop {L}\nolimits_{D}^n = \frac{1}{b}\sum_{x\in \mathop {\bx}\nolimits^{(r)}} {\log (\mathop D\nolimits_n  (x))} + \
\frac{1}{b}\sum_{x\in \mathop {\bx}\nolimits^{(d)}} {\log (1 - \mathop D\nolimits_n (x))},\\
&\mathop {L}\nolimits_{G}^n  = \frac{1}{b}\sum_{x\in \mathop {\bx}\nolimits^{(g)}}  {\log (1 - \mathop D\nolimits_n  (x))}.
\end{align}
\vspace{-0.15cm}
\end{subequations}
We present a more detailed description of FGAN-\uppercase\expandafter{\romannumeral2} in Algorithm 2.

In FGAN-\uppercase\expandafter{\romannumeral2}, instead of transmitting the synthesized data samples, we only perform coordination on the intermediate training parameters that are independent from the data dimension and training batch size. Unfortunately, both fog servers and the coordinator need to allocate more computation resources for the joint training process.
\vspace{-0.2cm}
\subsection{Global Service Classification}
\vspace{-0.2cm}
  In this step, we establish a global service classifier based on the synthetic samples produced by the generator at the global coordinator. Trained with high-quality synthesized global mixed service traffic data across all the decentralized local datasets, this classifier is able to identify different services at each fog server. Two procedures are needed to generate high accuracy global service classification: service sample labeling and global classifier training. We present a step-by-step description in Algorithm 3.

\subsubsection {Service Sample Labeling}
 In this procedure, we adopt a deep clustering method, referred to as DEC \cite{DEC}, to create a pseudo-label for each sample in the synthesized dataset, denoted as $T = \{ \hat {\mathop x}_1, \hat{\mathop x}_2, \dots, \hat{\mathop x}_{n_T} \}$.
In particular, we optimize a pre-trained  encoder and cluster centroid by minimizing the KL divergence between a target distribution  $p_{ij}$ and a soft assignment distribution  $q_{ij}$ as follows:
\vspace{-0.2cm}
\begin{equation} \small \label{12}
\begin{aligned}
L = KL(P||Q) = \sum\nolimits_i {\sum\nolimits_j {\mathop p\nolimits_{ij} \log \frac{{\mathop p\nolimits_{ij} }}{{\mathop q\nolimits_{ij} }}} }.
\end{aligned}
\vspace{-0.2cm}
\end{equation}
Let $Z_i$ be the latent feature of $\hat {\mathop x}_i$ extracted by the above mentioned encoder, and $\mathop \mu \nolimits_j $ be the centroid of cluster $j$. Then, $q_{ij}$ can be formulated as:
\vspace{-0.3cm}
\begin{equation} \small \label{10}
\begin{aligned}
{q_{ij}} = \frac{{{\rm{1 + (}}||{\rm{ }}{Z_i} - {\mu _j}|{|^2}{)^{{\rm{ - 1}}}}}}{{\sum\limits_{j = 1}^k {{{(1 + (||{\rm{ }}{Z_i} - {\mu _j}|{|^2}))}^{ - 1}}} }},
\end{aligned}
\vspace{-0.18cm}
\end{equation}
where $k$ is number of clusters.
In order to improve the cluster purity, the target distribution $p_{ij}$ is formulated as first squaring the soft assignment distribution and then normalizing it
\vspace{-0.2cm}
\begin{align}\label{11}
\mathop p\nolimits_{ij}  = \frac{{{{\mathop {\mathop q\nolimits_{ij} }\nolimits^2 } \mathord{\left/
 {\vphantom {{\mathop {\mathop q\nolimits_{ij} }\nolimits^2 } {\mathop f\nolimits_j }}} \right.
 \kern-\nulldelimiterspace} {\mathop f\nolimits_j }}}}{{{{\sum\nolimits_j {\mathop {\mathop q\nolimits_{ij} }\nolimits^2 } } \mathord{\left/
{\vphantom {{\sum\nolimits_j {\mathop {\mathop q\nolimits_{ij} }\nolimits^2 } } {\mathop f\nolimits_j }}} \right.
 \kern-\nulldelimiterspace} {\mathop f\nolimits_j }}}},
\vspace{-0.5cm}
\end{align}
where $\mathop f\nolimits_j  = \sum\nolimits_i {\mathop q\nolimits_{ij} } $ is the soft clustering probability. In this way, data points  $\{Z_1, Z_2, \cdot, Z_{n_T}\}$ can be assigned with a higher confidence. After the training process is completed, all the samples $\hat {\mathop x}_i$ in the synthesized dataset $T$ will be labeled via the optimal solution of maximizing $q_{ij}$.


To find the optimal number of clusters, Bayesian information criterion (BIC) is also adopted in our algorithm. For each possible cluster number $k$, we calculate a BIC value for clustering performance evaluation by:
\vspace{-0.2cm}
\begin{equation} \small \label{13}
\begin{aligned}
BIC_k = n_T \times \ln(\frac{R}{n_T }) + k \times \ln (n_T ),
\end{aligned}
\end{equation}
\vspace{-0.4cm}
\begin{equation} \small \label{14}
\begin{aligned}
R = \sum\limits_{j = 1}^k {\sum\limits_{\mathop Z\nolimits_i  \in \mathop \mu \nolimits_j } {\sqrt {\mathop {(\mathop Z\nolimits_i  - \mathop \mu \nolimits_j )}\nolimits^2 } } },
\end{aligned}
\vspace{-0.2cm}
\end{equation}
where $R$ is the sum of all the Euclidean distances between latent features $Z_i$ and its corresponding cluster centroid $\mu_j$, $n_T$ is the number of samples in the synthesized dataset $T$ and optimal number of cluster, denoted as $k^*$, will be found by first calculating BIC values with all possible $ k\in\{1, 2, ..., K_{max}\}$, then choosing the $k$ whose BIC value decreasing the most.

\begin{CJK}{UTF8}{gkai}
    \begin{algorithm}
\scriptsize
        \caption{Autonomous Service Classification}
            {\bf Input:} synthesized dataset $T = \{ \hat {\mathop x}_1 ,\hat{\mathop x}_2 ,\dots,\hat{\mathop x}_{n_T} \}$; maximum number of clustering $ K_{max}$; stopping threshold $\delta$; maximum iterations $I_{max}$;\\
   {\bf Output:} global classifier $C$;\\
      \quad {\bf for} $k = \{1, 2, ..., K_{max}\} $ {\bf do}
\begin{itemize}
      \item[] {\bf for} iteration round $i \leq I_{max}$ {\bf do}
      \item[] \quad 1)~initialize encoder and cluster centroid $\mu_j$;
      \item[] \quad 2)~mapping $\hat{\mathop x}_{i}$ to $Z_i$ with the encoder;
      \item[] \quad 3)~compute the distance between $Z_i$ and $\mu_j$ by equation (\ref{10});
      \item[] \quad 4)~label each $Z_i$ with the closest cluster index;
      \item[] \quad 5)~save labeled dataset as:
      \item[] \quad \quad \quad $T_{new} = \{ (\hat{x}_1,y_1) ,(\hat{x}_2,y_2),\dots,(\hat{x}_{n_T},y_{n_T}) \}$;
      \item[] \quad 6)~optimize encoder and $\mu$ according the clustering loss in equation (\ref{12});
      \item[] \quad 7)~re-calculate $y$ with the updated clustering model;
     \item[] \quad 8)~{\bf if} sum$(y_{old}\neq y)/n_T \le \delta $ {\bf then }
      \item[]  \quad \quad \quad Stop training;
    \item[] {\bf end for}
    \item[] Calculate ${BIC}_k$ value by equation (\ref{13});

    \item[] $\Delta BIC$=$BIC_k$-$BIC_{k-1}$;
\end{itemize}
   \quad {\bf end for} \\
    \quad  Find the optimal clustering model with maximum $\Delta BIC$ value; \\
    \quad  Calculate assignment probability $q_{ij}$; \\
     \quad  Label $\hat{x}_i$ with $\mathop{\arg\max}_{j} q_{ij}$;    \\
     \quad  Train service classifier $C$ with  $T_{new}$;\\
     \quad  Broadcast $C$ to all the fog servers;
    \end{algorithm}
\end{CJK}

\subsubsection {Global Classifier Training}
After autonomous service sample labeling, the original synthesized dataset $T$ will update to $T_{new} = \{ (\hat{x}_1,y_1) ,(\hat{x}_2,y_2),\dots,(\hat{x}_{n_T},y_{n_T}) \}$, where $y_i$ is the pseudo-label of $\hat{x}_i$. Then, this labeled  dataset will be utilized for training a global service  classifier $C$. When the model converges, this well-trained classifier will be broadcasted to all the connected fog servers for local service identification.
\vspace{-0.3cm}
\subsection{Autonomous Model Updating}
\vspace{-0.2cm}
In this step, we introduce a self-updating scheme to automatically update both the synthesizing and the classification model when new types of service traffic arrive. The proposed self-updating scheme consists of two procedures described as follows:
\subsubsection{Unknown Service Filtering}
For a given service traffic sample, the well-trained service classifier $C$ can output a vector indicating the probabilities of it belongs to the known service classes. We can monitor whether a new type of service emerges by periodically feeding a bulk of real service traffic data to the classifier $C$. Let $\bf o$ be the multi-dimensional output of classifier $C$ when feeding with a real service sample $s$, and we use $o^*=max(\bf o)$ to denote the confidence score of $s$. When $s$ belong to a  newly appeared service type for all the connected fog servers, $o^*$ will be relatively small, indicating  the well-trained classifier $C$ cannot assign this sample to any known service type with a high probability. Thus, we introduce a threshold value $\alpha$ to determine whether a traffic sample belongs to a known service type or not. That is, samples with $o^*<\alpha$ will be treated as unknown service types.
\subsubsection{Data Synthesizing and Classifier Updating}
Once the new type of service is identified in the above procedure, FGAN will be trained again with the updated datasets, which contain both the unknown service samples and the original dataset. In this way, a new mixture of traffic samples will be synthesized which capturing the updated services data distribution. Similarly, the newly synthesized dataset will first be labeled with pseudo-labels, and then used for classifier training. Finally, the re-trained classifier which is able to distinguish more diversified services will be broadcasted to each fog server for service classification again.

\vspace{-0.2cm}
\section{ Performance Evaluation}
  \vspace{-0.2cm}
\subsection{Dataset Description and Environmental Setup}
\vspace{-0.15cm}
  To evaluate the performance of our proposed framework on both service data synthesizing and classification, we conduct extensive simulations based on a real-world dataset  ``ISCX VPN-nonVPN dataset'' (ISCXVPN2016) which consisting of rich diversities of traffic samples including Emails, Facebook, SCP, etc. We consider data samples associated with 10 servers among which 8 services are considered as the known  traffic classes and 2  services are assumed to be unknown services as shown in Table \uppercase\expandafter{\romannumeral1}. In our simulations, data samples of both known and unknown services are uniformly randomly allocated to fog servers.

      All the simulations are conducted on a workstation with Intel(R) Core(TM) i5-8500 CPU@3.00GHz, 16.0 GB RAM@2133 MHz, 2 TB HD and four NVIDIA Corporation GP102 [TITAN X] GPUs and implemented with Pytorch library.
\vspace{-0.17in}
\begin{table}[htbp]
\scriptsize
  \centering
  \caption{Detailed Information of the Dataset for Evaluation}
  \vspace{-0.1in}
    \begin{tabular}{|c|c|c|}
\hline
          & \multicolumn{2}{|c|}{Application} \\
\hline
    \multirow{4}[0]{*}{ Existing classes}
& $Email$ & $Facebook$ \\
\cline{2-3}
          & $Netflix$ & $SFTP$ \\
\cline{2-3}
          & $Skype$ & $Vimeo$ \\
\cline{2-3}
          & $SCP$   & $Twitter$ \\
\hline
     Unknown classes & $Youtube^*$ & $VOIPbuster^*$ \\
\hline
    \end{tabular}%
  \label{tab:addlabel}%
  \vspace{-0.16in}
\end{table}%
\normalsize

\vspace{-0.2cm}
\subsection{Evaluation Results}
\vspace{-0.15cm}
To evaluate and compare the data synthesizing performance of FGAN-\uppercase\expandafter{\romannumeral1} and FGAN-\uppercase\expandafter{\romannumeral2}, we measure their communication overheads using the numbers of transmitted data bits and computational loads using the running time per global training round in Table \uppercase\expandafter{\romannumeral2} . We also use Maximum Mean Discrepancy (MMD) to evaluate the quality of samples synthesized by two FGAN methods which can be calculated by:
\vspace{-0.15cm}
\begin{equation} \small
\begin{aligned}
M M D^{2}\left(P_{r}, P_{g}\right)= \\\mathop{\mathbb{E}}\limits_{x_{r}, x_{r}^{\prime} \sim P_{r}, x_{g}, x_{g}^{\prime} \sim P_{g}}&\left [g\left(x_{r}, x_{r}^{\prime}\right)- 2 g\left(x_{r}, x_{g}\right)+g\left(x_{g}, x_{g}^{\prime}\right)\right],
 \end{aligned}
 \vspace{-0.15cm}
\end{equation}
where $x_r$ and $x_{r}^{\prime}$ are real samples drawn from real data distribution $P_{r}$, $x_g$ and $x_{g}^{\prime}$ are synthesized samples drawn from fake data distribution $P_{g}$.  $g(\cdot)$ represents a kernel function that maps the sample space to the Hilbert space. It can be observed that the communication overhead of FGAN-\uppercase\expandafter{\romannumeral1} is proportional to the batch size $b$, while the communication overhead of FGAN-\uppercase\expandafter{\romannumeral2} is mainly determined by the size of the model parameters. Compared to FGAN-\uppercase\expandafter{\romannumeral2}, FGAN-\uppercase\expandafter{\romannumeral1} consumes less running time per training round. With the number of training rounds increases, both FGAN-\uppercase\expandafter{\romannumeral1} and FGAN-\uppercase\expandafter{\romannumeral2} can obtain lower MMD scores which means that the quality of synthesized sample is improved.

\begin{table}[htbp]
  \centering
\tiny
  \caption{Evaluation of FGAN-I and FGAN-II}
  \vspace{-0.1in}
    \begin{tabular}{|c|c|c|c|c|c|c|c|c|}
\hline
    \multirow{2}[0]{*}{$I$} & \multirow{2}[0]{*}{$b$} & \multicolumn{1}{|c|}{\multirow{2}[0]{*}{$N $}} & \multicolumn{2}{|c|}{MMD} & \multicolumn{2}{|c|}{Com.overhead (MB)} & \multicolumn{2}{|c|}{Time per round (s)} \\
\cline{4-9}
          &       &       & FGAN-\uppercase\expandafter{\romannumeral1} & FGAN-\uppercase\expandafter{\romannumeral2} & FGAN-\uppercase\expandafter{\romannumeral1} & FGAN-\uppercase\expandafter{\romannumeral2} & FGAN-\uppercase\expandafter{\romannumeral1} & FGAN-\uppercase\expandafter{\romannumeral2} \\
\hline
    5     & 64    & 30    & 2.25  & 1.94  & 6.3  & 4.08 & 651.32 & 1571.77 \\
\hline
    10    & 64    & 30    & 0.67  & 1.45  & 6.3  & 4.08 & 651.32 & 1571.77 \\
\hline
    20    & 64    & 30    & 0.26  & 0.56  & 6.3  & 4.08 & 651.32 & 1571.77 \\
\hline
    5     & 128   & 30    & 0.96  & 0.96  & 12.5 & 4.08 & 347.41 & 1051.89 \\
\hline
    10    & 128   & 30    & 0.83  & 0.83  & 12.5 & 4.08 & 347.41 & 1051.89 \\
\hline
    20    & 128   & 30    & 0.35  & 0.38  & 12.5 & 4.08 & 347.41 & 1051.89 \\
\hline
    5     & 128   & 50    & 1.07  & 1.33  & 12.5 & 4.08 & 405.74 & 1288.84 \\
\hline
    10    & 128   & 50    & 0.79  & 0.62  & 12.5 & 4.08 & 405.74 & 1288.84 \\
\hline
    20    & 128   & 50    & 0.34  & 0.43  & 12.5 & 4.08 & 405.74 & 1288.84 \\
\hline
    \end{tabular}%
  \label{tab:addlabel}%
\vspace{-0.1in}
\end{table}%

\begin{table}[htbp]
  \centering
  \scriptsize
  \caption{Performance Evaluation Results Of The Classifiers}
  \vspace{-0.12in}
    \begin{tabular}{|c|c|c|c|c|}
    \hline
    \multicolumn{2}{|c|}{} &\textbf{Recall} & \textbf{Precision} & \textbf{F1 score} \\
    \hline
    \multirow{2}[0]{*}{\textbf{MLP}} & original & 0.3746 & 0.3787 & 0.3725 \\
     \cline{2-5}
          & updated & 0.3219 & 0.3273 & 0.3298 \\
    \hline
    \multirow{2}[0]{*}{\textbf{1D-CNN}} & original & 0.4128 & 0.4181 & 0.4135 \\
      \cline{2-5}
          & updated & 0.3767 & 0.3748 & 0.3654 \\
    \hline
    \multirow{2}[0]{*}{\textbf{2D-CNN}} & original & 0.4584 & 0.4564 & 0.4627 \\
     \cline{2-5}
          & updated & 0.4289 & 0.4263 & 0.4257 \\
    \hline
    \end{tabular}%
  \label{tab:addlabel}%
  \vspace{-0.17in}
\end{table}%
\normalsize

To evaluate the performance of classification , we consider  three evaluation metrics: Recall, model precision and F1 score, we consider two scenaries for each solution: (1) original: only 8 known services have been evaluated and classified, and (2) updated: total 10 services including 8 known and 2 unknown services have been evaluated. We use TP to denote the true positive decision that assigns two similar samples to the same cluster, and TN to denote the true negative decision that assigns two different samples to different clusters. We name the possible erroneous outputs of a classifier as false positive (FP) decision that assigns two different samples to the same cluster
and false negative (FN) decision that assigns two similar packets
to different clusters.
 We can then calculate those evaluation metrics as follows:
\vspace{-0.15cm}
\begin{equation} \small
\begin{aligned}
R = \frac{{TP}}{{TP + FN}} ,\quad P = \frac{{TP}}{{TP + FP}} ,\quad F1 = \frac{{2P \times R}}{{P + R}}
\end{aligned}
\vspace{-0.15cm}
\end{equation}
 We considere three network architectures for classifier: Multilayer Perceptron (MLP), 1D-CNN and 2D-CNN. Note that we resize the input data to 1 $\times$ 1600 for both MLP and 1D-CNN, and 40 $\times$ 40 for 2D-CNN. It can be observed that the 2D-CNN-based classifier offers the best performance. The MLP-based classifier has the lowest classification accuracy among three solutions which is around 37\%. We can also observe that compared to the original classifier, the performance of the updated classifier also degrades,  which may be attributed to two aspects: some unknown service samples are not detected, and some errors occured in the clustering process for the unknown service types, to slightly worsen the subsequent classification.
\vspace{-0.2 in}
\begin{figure}[htb]
\centering
\includegraphics[width=8.0cm]{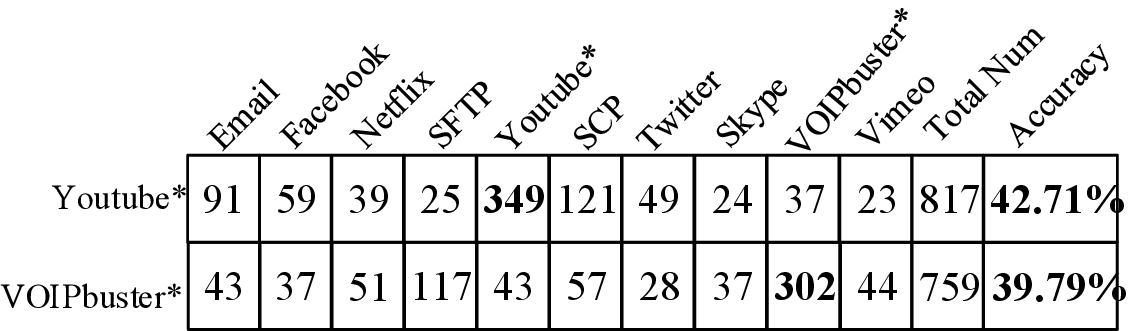}
\vspace{-0.15in}
\caption{Classification results for unknown classes(2D-CNN)}
\vspace{-0.15in}
\end{figure}

We present the classification result of the unknown services in Figure 3 which we present a confusion matrix to show the detailed classification results of two unknown service traffics. The confusion matrix calculates the classification accuracy by comparing the actual class of each data sample with the estimated one. We can observe that the accuracy for Youtube has reached roughly 42.71\%,  while the accuracy for VOIPbuster is around 39.79\%.
\vspace{-0.1in}
\section {Conclusion}
\vspace{-0.1in}
In this paper, we have proposed a novel framework that not only synthesizes the mixture of distribution of decentralized dataset but also classifies multiple unknown classes of data traffic. Furthermore, we have compared two types of FGAN approaches and propose an automatic traffic classification framework to autonomous classify unknown services and assign new pseudo-labels. Extensive simulations have been conducted and the numerical results demonstrate that our proposed framework is able to synthesize high-quality mixed service data traffic as well as significantly improve the service classification performance.
\vspace{-0.1in}
\section*{Acknowledgment}
This work was supported in part by the National Natural Science Foundation of China under Grants 62071193 and 61632019, the Key R \& D Program of Hubei Province of China under Grant 2020BAA002, and China Postdoctoral Science Foundation under Grant 2020M672357.

\bibliographystyle{IEEEtran}
\bibliography{ref}
\end{document}